\documentclass[pdflatex,sn-vancouver-num]{sn-jnl}% Vancouver Numbered Reference Style
\usepackage{graphicx}%
\usepackage{multirow}%
\usepackage{amsmath,amssymb,amsfonts}%
\usepackage{amsthm}%
\usepackage{mathrsfs}%
\usepackage[title]{appendix}%
\usepackage{xcolor}%
\usepackage{textcomp}%
\usepackage{manyfoot}%
\usepackage{booktabs}%
\usepackage{algorithm}%
\usepackage{algorithmicx}%
\usepackage{algpseudocode}%
\usepackage{listings}%
\usepackage{mdframed}
\theoremstyle{thmstyleone}%
\theoremstyle{thmstyletwo}%

\theoremstyle{thmstylethree}%

\hypersetup{
pdftitle={Behavioral Imaging Toolbox},
pdfsubject={cs.CV, q-bio.NC},
pdfauthor={Evangelos Sariyanidi, Birkan Tunc},
pdfkeywords={Computer Vision, AI, Psychology, Psychiatry},
}

\definecolor{code_bg}{RGB}{248, 243, 240}
\definecolor{code_fg}{RGB}{28, 29, 31}
\definecolor{code_keyword}{RGB}{233, 65, 89}
\definecolor{code_string}{RGB}{0, 152, 23}
\definecolor{code_comment}{RGB}{169, 146, 102}
\definecolor{code_number}{RGB}{169, 106, 192}
\definecolor{infoboxbg}{RGB}{240,235,232}
\definecolor{infoboxborder}{RGB}{180,180,180}

\lstdefinestyle{bitboxstyle}{
    backgroundcolor=\color{code_bg},
    basicstyle=\ttfamily\small\color{code_fg},
    keywordstyle=\color{code_keyword},
    stringstyle=\color{code_string},
    commentstyle=\color{code_comment},
    numberstyle=\tiny\color{gray},                    
    numbersep=5pt,
    frame=none,
    breakatwhitespace=true,         
    breaklines=true,
    breakindent=0pt,
    keepspaces=true,                 
    numbers=left,                 
    showspaces=false,                
    showstringspaces=false,
    showtabs=false,                  
    tabsize=2
}

\mdfdefinestyle{bitboxinfobox}{
  backgroundcolor=infoboxbg,
  linewidth=0.7pt,
  roundcorner=4pt,
  innerleftmargin=10pt,
  innerrightmargin=10pt,
  innertopmargin=8pt,
  innerbottommargin=8pt
}

\begin{document}

\title[Bitbox: Behavioral Imaging Toolbox for Computational Analysis of Behavior from Videos]{Bitbox: Behavioral Imaging Toolbox for Computational Analysis of Behavior from Videos}

%% \author{%
%%   $^1$
%%        \And
%%         $^1$
%%        \And
%%         $^1$
%%        \And
%%         $^1$
%%        \And
%%         $^1$
%%        \And
%%         $^2$
%%        \And
%%         
%%        \And
%%         $^{1,2}$
%%        \And
%%         $^3$
%%        \And
%%         $^{1,2}$
%%        \And
%%         $^{1,2}$
%%        \And
%%        $^1$, $^2$, $^3$University of Pittsburgh\\

\author[1]{\fnm{Evangelos} \sur{Sariyanidi}}\email{sariyanide@chop.edu}

\author[1]{\fnm{Gokul} \sur{Nair}}\email{nairg1@chop.edu}
%%\equalcont{These authors contributed equally to this work.}

\author[1]{\fnm{Lisa} \sur{Yankowitz}}\email{yankowitzl@chop.edu}

\author[1,2]{\fnm{Casey} \sur{J. Zampella}}\email{zampellac@chop.edu}

\author[1]{\fnm{Mohan} \sur{Kashyap Pargi}}\email{pargim@chop.edu}

\author[3]{\fnm{Aashvi} \sur{Manakiwala}}\email{aashvi@seas.upenn.edu}

\author[1]{\fnm{Maya} \sur{McNealis}}\email{mcnealism@chop.edu}

\author[1,2]{\fnm{John} \sur{D. Herrington}}\email{herringtonj@chop.edu}

\author[4]{\fnm{Jeffrey} \sur{Cohn}}\email{jeffcohn@pitt.edu}

\author[1,5]{\fnm{Robert} \sur{T. Schultz}}\email{schultzrt@chop.edu}

\author*[1,2,6,7]{\fnm{Birkan} \sur{Tunc}}\email{tuncb@chop.edu}

\affil[1]{\orgdiv{Center for Autism Research}, \orgname{The Children's Hospital of Philadelphia}, \city{Philadelphia}, \postcode{19146}, \state{PA}, \country{USA}}

\affil[2]{\orgdiv{Department of Psychiatry}, \orgname{University of Pennsylvania}, \city{Philadelphia}, \postcode{19104}, \state{PA}, \country{USA}}

\affil[3]{\orgdiv{Department of Computer and Information Science}, \orgname{University of Pennsylvania}, \city{Philadelphia}, \postcode{19104}, \state{PA}, \country{USA}}

\affil[4]{\orgdiv{Department of Psychology}, \orgname{University of Pittsburgh}, \city{Pittsburgh}, \postcode{15260}, \state{PA}, \country{USA}}

\affil[5]{\orgdiv{Department of Pediatrics}, \orgname{University of Pennsylvania}, \city{Philadelphia}, \postcode{19104}, \state{PA}, \country{USA}}

\affil[6]{\orgdiv{Department of Biomedical and Health Informatics}, \orgname{The Children's Hospital of Philadelphia}, \city{Philadelphia}, \postcode{19146}, \state{PA}, \country{USA}}

\affil[7]{\orgdiv{Center for Biomedical Image Computing and Analytics (CBICA)}, \orgname{University of Pennsylvania}, \city{Philadelphia}, \postcode{19104}, \state{PA}, \country{USA}}

\abstract{
Computational measurement of human behavior from video has recently become feasible due to major advances in artificial intelligence. These advances now enable granular and precise quantification of facial expression, head movement, body action, speech behavior, and other behavioral modalities and are increasingly used in psychology, psychiatry, neuroscience, and mental health research. However, mainstream adoption remains slow. Most existing methods and software are developed for engineering audiences, require specialized software stacks, and fail to provide behavioral measurements at a level directly useful for hypothesis-driven research. As a result, there is a large barrier to entry for researchers who wish to use modern, AI-based tools in their work. We introduce Bitbox, an open-source toolkit designed to remove this barrier and make advanced computational analysis directly usable by behavioral scientists and clinical researchers. Bitbox is guided by principles of reproducibility, modularity, and interpretability. It provides a standardized interface for extracting high-level behavioral measurements from video, leveraging multiple face, head, and body processors. The core modules have been tested and validated on clinical samples and are designed so that new measures can be added with minimal effort. Bitbox is intended to serve both sides of the translational gap. It gives behavioral researchers access to robust, high-level behavioral metrics without requiring engineering expertise, and it provides computer scientists a practical mechanism for disseminating methods to domains where their impact is most needed. We expect that Bitbox will accelerate integration of computational behavioral measurement into behavioral, clinical, and mental health research. Bitbox has been designed from the beginning as a community-driven effort that will evolve through contributions from both method developers and domain scientists.
}

%%\keywords{keyword1, Keyword2, Keyword3, Keyword4}



\maketitle

\section{Introduction}\label{sec1}
Understanding human behavior and its variation is central across psychology, sociology, cognitive science, economics, politics, and many other fields. Progress in all these domains depends on how accurately, granularly, and richly we capture and represent behavior in natural contexts. This need becomes even more critical in clinical psychology, psychiatry, and mental health, in which conditions are defined mainly through observable behavior—what a person does well or poorly, too little or too much. In this context, better behavioral measurement is not merely desirable: it is fundamental for improving diagnosis, prognosis, intervention design, and treatment monitoring. 

In the last few decades, we have witnessed rapid development of biological methods, such as genomic sequencing and neuroimaging, that provide rich representations of biological mechanisms. On the other hand, methods to measure the observable expression of these mechanisms—the behavioral phenotype—have, until now, been stuck in time, creating an “information gap” between biology and behavior. Rich biological signals are often mapped to broad symptom scores or, worse, binary case–control labels, a mismatch that has contributed to underpowered, nonreplicable, and nongeneralizable findings \citep{Krakauer2017}. This information gap is detrimental for efforts to identify biological mechanisms underlying behavior and its impairments, such as brain circuits and gene expression networks. 

Traditional approaches for measuring behavior, including self-reports, expert ratings, and performance-based tasks, contribute to this information gap, as they fall short on ecological validity, reliability, scalability, precision, and granularity. Self-reports are influenced by confounds \citep{Rutter2003} and often reflect broad traits rather than specific behaviors \citep{Anagnostou2015}, with poor inter-rater agreement \citep{Stratis2015,Mitsis2000,Azad2016,Kaat2015,Achenbach2005}. Expert assessments are time-intensive, provide brief behavioral snapshots \citep{Paul2009}, and often show lower than desired reliability \citep{Regier2013}.

Fine-grained annotation systems such as the Facial Action Coding System (FACS) \citep{EkmanFriesen1978} and the Body Action and Posture (BAP) coding system \citep{Dael2012} can, in principle, extract richer behavioral signals, but they require time-consuming coding by trained experts, drastically limiting their applicability for measuring natural behavior in real-world applications. There is a pressing need for scalable approaches that collect rich behavioral samples with minimal burden, enabling low-cost, high-yield, real-world “big data” studies, especially as telemedicine and mobile health expand.

Novel experimental paradigms and technologies are now emerging and changing that landscape \citep{Hamilton2023} within a Computational Behavior Analysis (CBA) paradigm. Using scalable sensors (e.g., phone cameras, wearables) with computer vision, affective computing, and language technologies, CBA quantifies facial expressions, head and body movements, gaze, speech, and interpersonal coordination with precision and consistency, even from brief, naturalistic recordings \citep{Zampella2021,Sariyanidi2023workshop,ParishMorris2018,Gokmen2024}. It is now feasible to measure spontaneous, multimodal, and natural behavior as it unfolds in socially relevant settings—a key need across psychology, medicine, and the social sciences \citep{Mondada2016,HollerLevinson2019,Diana2023,Gratch2023,Freeth2023,Girard2026ARCP}.

Current applications of CBA chiefly span developmental, social, and clinical psychology; affective and cognitive science; and social neuroscience \citep{Goldstein2020,Zhao2022,Wohltjen2021,Hirsch2023,Cheong2023,Masip2014,Haines2019,Martinez2019,Barrett2019,Martin2021,Shen2022,Neuman2023,Peterson2022,Bilalpur2023ICMI,Hinduja2024TAFFC}. Early clinical deployments show promise across screening, diagnosis, and treatment monitoring \citep{Koutsouleris2022,Perochon2023}. For example, recent multiclinic studies leveraged computer vision and eye-tracking to detect subtle social behaviors for early screening of autism during routine care \citep{Jones2023}. These approaches offer reproducibility, scalability, and sensitivity to subtle variation, positioning them to transform behavioral and medical science.

Despite this promise, mainstream adoption remains slow. CBA integrates methods from computer science, robotics, and bioengineering with application needs across behavioral, social, and medical domains. Teams must navigate this multidisciplinary space while most existing tools are built by/for engineers, depend on specialized tech stacks, and rarely expose high-level, interpretable behavioral measurements directly usable for hypothesis-driven research.

In this article, we introduce the Behavioral Imaging Toolbox (Bitbox)~\citep{BitboxDocs_Intro}, an open-source Python library designed to bridge engineering innovation and behavioral-science application\footnote{Source code can be accessed at https://github.com/compsygroup/bitbox}. Bitbox provides standardized, validated measures of nonverbal behavior. Nonverbal behavior, including facial expression, head movement, eye gaze, body action, speech behavior, and prosody, plays a central role in communication, emotion regulation, and mental health. %%Social interaction depends on the perception and production of coordinated face–body signals \citep{Gu2013}. The human face and eyes are among the richest sources of social information \citep{Jack2017}, conveying cues about personality \citep{Krumhuber2007}, mental states \citep{Nusseck2008}, and emotion \citep{Ekman1969}. Head and body movements and posture provide additional channels for information transfer during social communication \citep{deGelder2009,Frijda1988,Roether2009}; even the movement of a single limb can alter how we perceive another’s emotional experience \citep{Pollick2001}. Impairments in nonverbal behaviors are widespread across psychiatric and neurodevelopmental conditions such as anxiety and mood disorders \citep{Clark1991,Ellgring1989,Girard2014,Varlet2014}, autism \citep{Howlin2017,Shaw2025}, ADHD \citep{Fliers2008,FenollarCortes2017} and psychosis \citep{Scheffer2004}. 
Bitbox prioritizes usability for non-engineers yet retains power and flexibility for affective computing and AI research. As explained in its online documentation~\citep{BitboxDocs_Intro}, installation and use are straightforward and require no engineering skills. By unifying diverse computer-vision backends behind a common API, Bitbox enables tool-agnostic behavioral constructs, and supports interactive visualization. At the same time, it provides computer scientists a practical mechanism for disseminating their sophisticated methods to application domains where their impact is most needed.

In the following sections, we discuss Bitbox's relation to existing software, outline its system architecture, describe the programming interface with an example workflow, demonstrate its visualization capabilities, and detail open-science and reproducibility features. Together, these elements present a practical, reproducible pathway for adopting CBA in behavioral research. Bitbox lowers the barrier to high-quality behavioral measurement at scale, to make advanced tools usable in everyday research practice.

\section{Relation to Existing Software}

The past decade has seen a rapid expansion of computer vision, natural language processing, and signal-processing tools capable of quantifying human motor, social, emotional, and affective behavior. With advances in machine learning and deep neural networks, these tools now reach impressive performance on large public benchmark datasets and are gradually becoming accessible to behavioral and clinical researchers. This technological momentum has opened the possibility of measuring everyday human behavior from video, audio, and text at a scale that was unimaginable only a few years ago.

A wide range of tools are available for different behavioral modalities. For facial expression analysis, commonly used systems include MediaPipe~\citep{Lugaresi2019MediaPipe}, OpenFace~\citep{Baltrusaitis2018OpenFace}, PyAFAR~\citep{Hinduja2023PyAFAR}, Py-Feat~\citep{Cheong2023PyFeat}, LibreFace~\citep{Chang2024LibreFace}, 3DI~\citep{Sariyanidi2024TPAMI}, and 3DDFA-V2~\citep{Guo2020ECCV3DDFAv2}. For gaze analysis, tools such as MediaPipe, OpenFace, RT-GEN~\citep{Fischer2018RTGENE}, and EyeGestures~\citep{Niehorster2025EyeGestures} provide pupil or head-aligned gaze estimates. Head-movement tracking can be performed with MediaPipe, OpenFace, 3DI, OpenPose~\citep{Cao2017OpenPose}, or AlphaPose~\citep{Fang2022AlphaPose}. Full-body motion can be extracted with OpenPose, AlphaPose, DeepLabCut~\citep{Mathis2018DeepLabCut}, Detectron~\citep{Girshick2018Detectron}, and MMPose~\citep{MMPose2020}. In parallel, speech-behavior tools such as OpenSMILE~\citep{Eyben2010openSMILE}, TalkNET~\citep{Tao2021TalkNet} and ASDNet~\citep{Kopuklu2021ASDNet} have begun to quantify prosody, turn-taking, and vocal markers. These tools demonstrate the breadth of engineering innovations relevant to behavioral science.

The translation of these methods to behavioral and clinical research, however, has been uneven. In practice, only a small subset of tools—most notably OpenFace~\citep{Baltrusaitis2018OpenFace} for facial analysis and OpenPose~\citep{Cao2017OpenPose} for body analysis—have gained widespread adoption among researchers. Their popularity reflects accessibility and ease of use rather than reliability, validity, or comprehensive behavioral coverage. Most available tools provide only low-level signals, such as action-unit activations, head-pose parameters, or skeletal keypoints. They leave out higher-order behavioral properties such as biomechanical characteristics, psychomotor profiles, expressive features, or social dynamics—concepts that map more directly onto theoretical constructs in psychology, psychiatry, and neuroscience.

Bitbox addresses this translational gap by providing an architecture (see Section~\ref{sec:architecture}) that incorporates state-of-the-art tools as backend processors, standardizes their outputs, and adds a dedicated measurement layer (see Table~\ref{tab:bitbox_measurements}) designed specifically for behavioral research. Instead of competing with existing computer vision methods, Bitbox uses them as signal generators and builds scientifically meaningful metrics on top. In this way, Bitbox augments rather than replacing existing tools, transforming their outputs into variables that can be directly integrated into statistical models, mixed-effects analyses, machine-learning frameworks, or any clinical research pipelines.

Two existing software platforms bear partial similarity to Bitbox, namely Py-Feat~\citep{Cheong2023PyFeat} and OpenSense~\citep{StefanovICMI2020}. Py-Feat unifies multiple action unit (AU) and emotion-detection tools under a common interface, but its output remains limited to AUs and emotion categories. OpenSense integrates data-acquisition and processing tools, providing a unified interface for generating low-level behavioral signals such as AUs, head pose, and gestures. Neither system provides high-level behavioral measurements tailored for psychological or clinical inquiry. Bitbox fills a unique space by offering both standardized backend integration and a library of domain-relevant behavioral constructs.

A key innovation of Bitbox is that its measurements were developed through close collaboration between engineering, behavioral science, and clinical research. The toolbox reflects not only state-of-the-art computational methods but also the conceptual frameworks and practical constraints of real-world behavioral studies. In this sense, Bitbox represents a synthesis of engineering advances with behavioral and clinical domain knowledge, making modern computer-vision tools genuinely useful for the scientific study of human behavior.

\section{System Architecture}
\label{sec:architecture}
Bitbox is designed to measure nonverbal behavior from video recordings. At present, it supports the quantification of face and head movement, with extensions for full-body actions, speech behavior (e.g., identifying who is speaking and when), and prosody under active development.

\begin{figure}[h]
    \centering
    \includegraphics[width=1.0\linewidth]{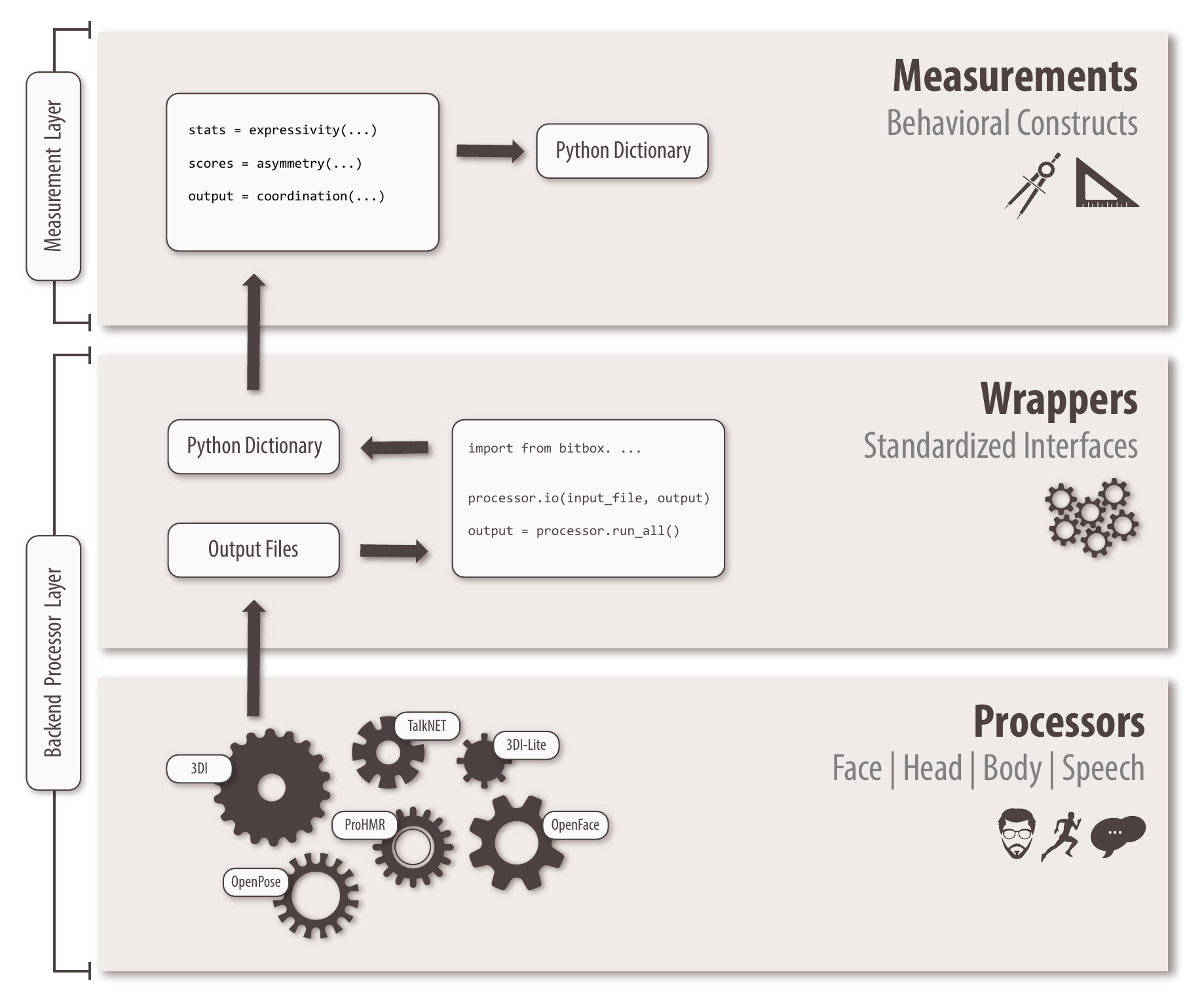}
    \caption{Bitbox system architecture, consisting of two main layers: (1) Backend processor layer, including standalone computer vision tools and associated wrapper functions and (2) measurement layer, including functions that derives high-level behavioral measurements from the outputs of processors. Note that current backends are limited to video processing. \tiny{Free graphics from Vexels are used in this illustration}} 
    \label{fig:system_architecture}
\end{figure}

Research using CBA typically follows a staged workflow. First, behavioral signals—such as facial expressions, gaze patterns, or head movements—are extracted from raw video data. These time-varying signals are then analyzed using machine learning models or computational algorithms to produce higher-level outcomes such as categorical labels (e.g., diagnostic or emotional classifications), predictive scores (e.g., risk for depression), or quantitative behavioral indices (e.g., engagement) ~\citep{cowan2024computerized,Mahmood2025SciRep,hou2021markerless,Lee2022SensorsADHD,fuzi2023emerging,Liu2025JAD,sonawane2021review,Ozturk2024PsycholMed,simmatis2022reliability,Hinduja2024TAFFC,ambrosen2025using,Muszynski2020ICMI,Sadeghi2024NPJMH,Celiktutan2023ICPRW,Sahu2025ArXivAnxietyFaceTrack,zhu2022review,Koehler2024TransPsychiatry,Gokmen2024,Zampella2021,Sariyanidi2023workshop}.

Bitbox implements this workflow through a two-layer architecture as illustrated in Fig~\ref{fig:system_architecture}. The lower layer consists of \emph{backend processors} that extract raw behavioral signals from video, frame by frame. These processors output features such as face bounding boxes, 2D and 3D facial landmarks, head pose, and expression coefficients. The upper layer organizes and standardizes these outputs, then computes higher-level behavioral measures derived from them using \emph{measurement routines}. These measures can be directly tested statistically or integrated into machine learning and AI models for predictive or mechanistic analyses.

\subsection{Backend Processors}
In Bitbox, a backend refers to any processing engine that receives raw video/audio input and produces time-varying behavioral signals. Examples may include facial landmark extractors that identify key points on the face (e.g., eye corners, nose tip, mouth corners) \citep{Bulat2017ICCVFaceAlignment,Zhu2019TPAMI3DDFA}, models that generate expression coefficients describing muscle movements across time \citep{Sariyanidi2024TPAMI}, and head-pose estimators that recover 3D orientation from 2D video frames \citep{Hempel2024pose}.

Rather than re-implementing these algorithms, Bitbox integrates third-party backends as they were originally developed and optimized by their creators. Each backend runs as a standalone executable that produces backend-specific output files. This design allows Bitbox to seamlessly adopt newer, state-of-the-art models as they become available, without requiring structural changes to the system.

Extracting meaningful behavioral measures often requires combining several processors even within a single modality. For example, analysis of facial expressions typically involves a sequence of steps: detecting and tracking the face region in each frame, identifying facial landmarks, estimating head pose, and computing expression coefficients that capture dynamic muscle activity over time. Together, these steps generate a detailed multivariate signal that describe how facial features move and interact.

However, many of these tools were developed by and for engineering audiences, often released as research prototypes in computer vision or affective computing venues. Their installation, configuration, and maintenance can be complex and time-consuming for behavioral or clinical researchers. To overcome this barrier, Bitbox provides prebuilt Docker images that encapsulate all dependencies and processors, eliminating the need for per-tool installation or environment setup. Users can therefore focus on scientific questions rather than software engineering.

\begin{figure}[h]
    \centering
    \includegraphics[width=1.0\linewidth]{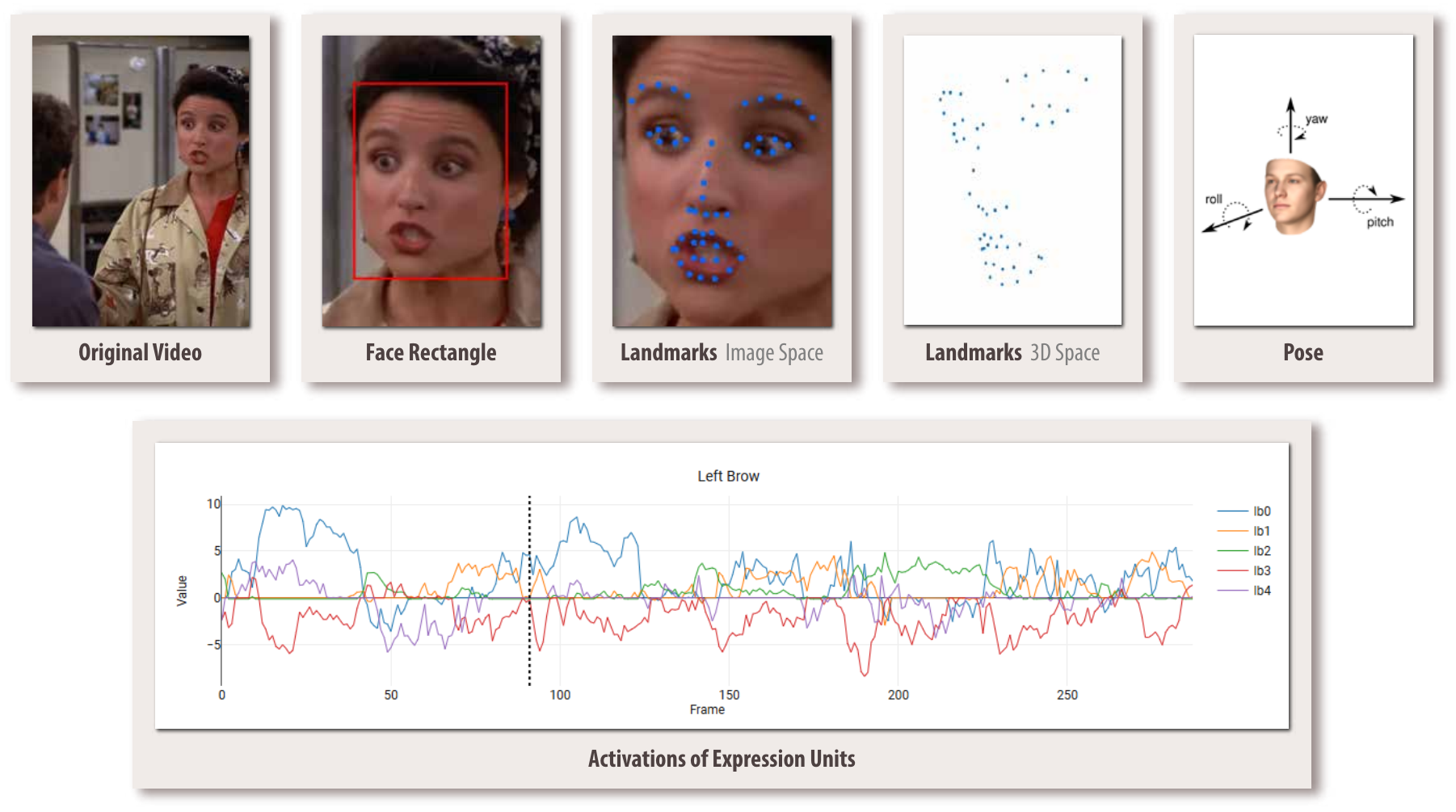}
    \caption{The most commonly used components of facial analysis workflows, including facial rectangles, landmarks (both 2D and 3D), pose, and expression related movements. Most facial analysis software provides these components, or use them internally.} 
    \label{fig:facial_components}
\end{figure}

Currently, Bitbox includes two facial processing backends, namely 3DI \citep{Sariyanidi2024TPAMI} and 3DI-Lite \citep{Sariyanidi2025}, and is actively expanding to include OpenFace~\citep{Baltrusaitis2018OpenFace}, as well as body-processing backends for full-body pose and limb kinematics \citep{chen2025human3r, yang2025sam3dbody}. These processors can generate all components necessary for head and facial movement analysis, including face rectangles, 2D landmarks, 3D canonicalized landmarks (corrected for head pose and individual morphology), head pose estimates, 3D facial meshes, and coefficients representing both morphology and expressions. See Fig.~\ref{fig:facial_components} for an illustration of these components. Internally, they also rely on several state-of-the-art third-party tools and pretrained models \citep{Bulat2017ICCVFaceAlignment,Paysan2009AVSS}.

Bitbox communicates with its backends through two lightweight wrapper functions: One handles execution of the processor, and the other reads and structures the processor outputs. All results are automatically converted into standardized dictionary formats that can be used consistently across the software. Once a backend is wrapped, it becomes immediately available throughout Bitbox, fully interoperable with all analysis modules.
Finally, the measurement layer of Bitbox—described in the next section—is intentionally designed to operate independently of specific backends. Measurements are derived from standardized, typed outputs rather than raw, backend-specific files. This modularity minimizes lock-in, allows users to test and compare different processors for the same task, and ensures that Bitbox can evolve alongside rapid advances in computer vision technology.

\subsection{Behavioral Measurements}
The raw outputs generated by backend processors capture biological movements, such as facial muscle activations, head rotations, and body trajectories, but these signals do not, on their own, map directly onto the behavioral constructs that matter in psychology, neuroscience, psychiatry, or social science. For example, time series of facial action units describe how specific muscles move, yet researchers often care about higher-order properties such as the \emph{symmetry} of expressive behavior, the \emph{overall intensity} or \emph{variability} of expressions, or how facial movements reflect affective regulation. Similarly, head-movement outputs may encode pitch, roll, and yaw, whereas a psychological study might instead need derived indicators of social attention, avoidance, orienting behavior, or motor control. When two people interact, combining signals across individuals is essential for estimating \emph{coordination}, \emph{synchrony}, or \emph{imitation}—constructs central to social, developmental, and clinical research.

To bridge this gap between low-level signals and theoretically meaningful constructs, Bitbox provides a dedicated \emph{behavioral measurement layer}. Each measurement routine receives one or more standardized backend outputs and computes interpretable variables—scores and indices—that can be used directly in statistical models or integrated into ML/AI pipelines. These measurements fall into three broad categories: \emph{biomechanical characteristics and psychomotor function}, \emph{affective expressions}, and \emph{social dynamics}. Table~\ref{tab:bitbox_measurements} lists all currently supported measurements; below we summarize the main groups and their relevance.

\begin{itemize}
    \item \textbf{Biomechanics.}  
    These measurements quantify movement structure and motor quality, including displacement amplitude, velocity, smoothness, tremor-like fluctuations, and range of motion. For facial and head movements, such biomechanical indices are important in research on psychomotor function, motor development, neurological soft signs, coordination difficulties in neurodevelopmental conditions, and links between motor behavior, genetics, and neural circuitry.

    \item \textbf{Affective expressions.}  
    Bitbox includes measures of expressivity (temporal variability in expression coefficients), expression intensity, symmetry, and diversity. These higher-level affective descriptors are widely used in studies of emotion regulation, mood and anxiety disorders, interpersonal communication, and affective neuroscience, where subtle differences in expressive behavior carry diagnostic and functional significance.

    \item \textbf{Social dynamics.}  
    These measurements quantify interpersonal processes such as imitation, synchrony, temporal alignment, and coordination between interacting partners. For example, correlations between head-pose trajectories can capture moment-to-moment coupling within dyads. Such indices are central in developmental, social, and clinical psychology (e.g., autism research); relationship science; and human–computer interaction.
\end{itemize}

\begin{table}[h]
\centering
\caption{Measurement routines included in Bitbox. Functions take the outcome of backend processors, such as facial landmarks or expression activations, and generate listed measurements.}
\label{tab:bitbox_measurements}
\renewcommand{\arraystretch}{1.3}
\begin{tabular}{p{3.2cm} p{9cm}}
\hline
\textbf{Name} & \textbf{Description} \\
\hline
\multicolumn{2}{c}{\textbf{Biomechanics}} \\
Range of motion 
& Computes the span between minimum and maximum positions over time, capturing movement extent. \\
Total path length 
& Sums the cumulative trajectory length of movement to quantify total movement. \\
Average speed 
& Computes the mean velocity of motion across frames. \\
Average acceleration 
& Calculates the mean rate of change in velocity over time. \\
Average jerk 
& Measures the average rate of change in acceleration, indexing movement smoothness or jerkiness. \\
Log dimensionless jerk
& Computes a normalized log‐scale jerk metric enabling comparison across individuals or time spans. \\
Relative motion 
& Measures displacement relative to a reference frame to characterize coordinated or differential motion. \\
\hline
\multicolumn{2}{c}{\textbf{Affective Expressions}} \\
Symmetry 
& Assesses left–right balance in facial movements to quantify facial asymmetry. \\
Expressivity 
& Quantifies the intensity and temporal variability of expressions. \\
Diversity 
& Measures the variability and range of expression types exhibited over time. \\
\hline
\multicolumn{2}{c}{\textbf{Social Dynamics}} \\
Imitation 
& Computes how closely one person’s behavior mirrors another’s, quantifying mimicry (unidirectional). \\
Coordination 
& Assesses temporal alignment or coupling between interacting partners’ behavior (bidirectional). \\
\hline
\end{tabular}
\end{table}

Thanks to Bitbox’s modular architecture and standardized output structures, adding new measurement routines is straightforward. Developers can extend the system by specifying the required backend outputs and implementing the computation logic; Bitbox manages input handling, metadata, and consistency. As the field evolves and new constructs emerge, this framework allows the community to expand Bitbox organically, ensuring that the toolbox remains aligned with scientific needs across many disciplines.

\begin{mdframed}[style=bitboxinfobox]
\textbf{Why Bitbox Does Not Include Emotion Labels} \\[4pt]
Many commercial and academic tools attempt to infer categorical emotion labels (e.g., sadness, anger, surprise) by mapping facial movements or other visual signals onto these categories~\citep{CowenKeltner2017PNAS,Schurr2024NatHumBehav}. Bitbox intentionally does not provide such categorical labels because the scientific foundations of third-person emotion labeling from video remain highly contested~\citep{Barrett2019PSPI,Jack2012PNAS}. A growing literature shows that predicting whether a person is ``happy'' or ``sad'' purely from images—without access to the individual’s context, culture, personal history, goals, or the surrounding interaction—is fundamentally ill-posed~\citep{Brooks2024iScience}. In its current form, Bitbox does not include behavioral modalities that would be needed for first-person affect estimation (e.g., language content, acoustic properties of speech, or physiological signals). Given these limitations, we believe the most scientifically responsible approach is to focus on describing observable movements and their kinematic, dynamic, morphological, and interpersonal characteristics, rather than predicting broad labels of internal emotional states.
\end{mdframed}

\section{Programming Interface}
Bitbox is a Python library, and users interact with it in the same way they would use any scientific computing package such as \texttt{scikit-learn}, \texttt{numpy}, or \texttt{pandas}. The workflow is straightforward: A user selects a backend processor, specifies input and output paths, runs the processor to extract behavioral signals, and then applies one or more measurement functions to convert these signals into interpretable behavioral scores. Although backend processors generate outputs in their own proprietary formats, Bitbox automatically reads these files and returns all results in a standardized Python dictionary structure. This allows analysis routines to work seamlessly across different tools and modalities. A typical workflow that generates face rectangles, landmarks, and expression coefficients and computes a behavioral variable (e.g., overall expressivity of face) is listed in Code~\ref{code:bitbox_workflow}.

\begin{lstlisting}[style=bitboxstyle, language=Python,
caption={Example Bitbox workflow to quantify overall expressivity of face},
label={code:bitbox_workflow}]
from bitbox.face_backend import FaceProcessor3DI as FP
from bitbox.expressions import expressivity

# define a face processor using the Docker image
processor = FP(runtime='bitbox:latest')

# define input file and output directory
input_file = 'data/elaine.mp4'
output_dir = 'output'

# set input and output
processor.io(input_file=input_file, output_dir=output_dir)

# detect faces
rects = processor.detect_faces()

# detect landmarks
lands2D = processor.detect_landmarks()

# estimate facial expressions, pose,
# and canonicalized landmarks in 3D
exp_global, pose, lands3D = processor.fit(normalize=True)

# compute expressivity stats
expressivity_stats = expressivity(exp_global, scales=6)
\end{lstlisting}

\subsection{Selecting and Using Processing Backends}
Bitbox supports multiple facial and (soon) full-body processing backends in a plug-and-play manner. Any backend can be integrated as long as the necessary wrapper functions to run the executable and to read its outputs are defined. This design allows researchers to choose the processor that best fits their scientific goals while maintaining a consistent downstream interface for measurement and analysis.

At present, Bitbox provides two facial analysis backends: \emph{3DI}~\cite{Sariyanidi2024TPAMI} and \emph{3DI-Lite}~\cite{Sariyanidi2025}. Both extract facial landmarks, head pose, and expression coefficients, but they differ in complexity and intended use. 3DI is a full, high-fidelity 3D facial analysis system that reconstructs a dense facial mesh and operates entirely in 3D space. It was designed for accurate behavioral analysis, research-grade modeling, and studies requiring stable, interpretable facial representations. 3DI-Lite is a lightweight version optimized for speed and scalability, as well as robustness to occlusions and partial visibility of the face when regions fall outside the camera frame (e.g., forehead cut off when someone is too close to the camera). It provides the core outputs needed for most behavioral research while offering significantly faster runtime, lower computational cost, and higher robustness to unexpected image quality issues.

The primary backend in Bitbox is \emph{3DI}, selected because of a crucial property: the separation of pose, expression, and identity~\cite{Sariyanidi2020CVPR}. Since 3DI reconstructs a 3D facial mesh, it can disentangle these components explicitly, producing coefficients for head pose, identity-related morphology, and stable facial expressions that are robust to pose and identity. It also allows Bitbox to generate \emph{canonicalized} 3D facial landmarks that are corrected for pose and identity, isolating the movement components that correspond purely to expression dynamics.

Using a backend in Bitbox follows a simple pattern. A researcher begins by importing the processor of interest and specifies the runtime environment—either a Docker image or a native on-machine installation of the backend.

\begin{lstlisting}[style=bitboxstyle, language=Python, label={lst:import_backend}]
# 3DI
from bitbox.face_backend import FaceProcessor3DI as FP
# 3DI-Lite
from bitbox.face_backend import FaceProcessor3DIlite as FP

# using a Docker image tagged as 'bitbox:latest'
processor = FP(runtime='bitbox:latest')

# using a native installation
processor = FP(runtime='/path/to/backend/bin')
\end{lstlisting}

To avoid specifying the runtime each time, users may instead set system-level environment variables~\cite{BitboxDocs_Intro}. After selecting a backend and choosing a runtime, the final step is to set input/output paths and run the processor. Bitbox handles execution and managing the backend’s proprietary outputs.
\begin{lstlisting}[style=bitboxstyle, language=Python]
processor.io(input_file='file.mp4', output_dir='out/')
combined_results = processor.run_all(normalize=True)
\end{lstlisting}

This unified interface ensures that even though backends may differ in complexity, file formats, or underlying algorithms, their outputs can be used interchangeably within the Bitbox measurement layer.

\begin{mdframed}[style=bitboxinfobox]
\textbf{Pose-Expression Ambiguity} \\[4pt]
Many behavioral and clinical studies rely on expression-related variation. Analyzing facial behavior in video involves quantifying expression from 2D frames, which can be challenging. Facial pose, expressions, and individual identity are intertwined in 2D space, complicating analysis~\cite{Sariyanidi2020CVPR}. For instance, a person viewed frontally may appear to frown if their head tilts downward, altering the distance between facial features. Furthermore, each person’s unique facial structure affects analysis, as variations in feature shapes and distances can skew expression detection. For example, naturally low eyebrows might also falsely suggest a frown.

The use of 3D modeling offers a solution by accurately fitting a 3D morphable model (3DMM) to the subject's face~\cite{Sariyanidi2024TPAMI}. This approach effectively disentangles pose, expression, and identity, providing expression coefficients that represent expressions free from interference by pose or identity traits. Bitbox's main face processor, 3DI, effectively removes these confounds and enables cleaner estimation of true expressive behavior. 
\end{mdframed}

\subsection{Running Measurement Functions}
Once backend processors have extracted facial landmarks, head pose, or expression coefficients, the next step is to derive higher-level measurements that map these low-level signals onto behavioral constructs relevant for psychological and clinical research. Bitbox provides a set of analysis functions, listed in Table~\ref{tab:bitbox_measurements}, that operate on the standardized backend outputs and return interpretable variables suited for statistical modeling or machine-learning workflows.

For biomechanical measurements, Bitbox allows users to compute kinematic characteristics from face rectangles, head pose, or facial landmarks (2D or 3D).

\begin{lstlisting}[style=bitboxstyle, language=Python]
# motion kinematics from face rectangles
mrange, path, speed, acc = motion_kinematics(rects)

# kinematics from head pose (using translation components)
mrange, path, speed, acc = motion_kinematics(pose, angular=False)

# kinematics from 3D landmarks
mrange, path, speed, acc = motion_kinematics(lands3D)
\end{lstlisting}

For face rectangles, Bitbox follows the motion of the rectangle center through time. For head pose, translational (\texttt{angular=False}) or rotational (\texttt{angular=True}) components yield 3D movement trajectories. When landmarks are used, each point is tracked independently in 2D or 3D. In most cases, 3D landmarks provide more stable and accurate measurements, since 2D landmarks are affected by head motion and camera perspective. Unless faces remain frontal throughout most frames, 2D landmarks should be avoided.

Symmetry is computed using facial landmarks. Because symmetry is defined as the disparity between the left and right halves of the face, obtained by mirroring each side across the perpendicular-bisector plane, 2D landmarks may produce invalid values when the face is not frontal. For this reason, 3D landmarks are strongly recommended.

\begin{lstlisting}[style=bitboxstyle, language=Python]
# per-frame asymmetry scores
sym3D = asymmetry(lands3D)

# use with caution unless face is consistently frontal
sym2D = asymmetry(lands2D)
\end{lstlisting}

Other affective-expression measurements rely on expression coefficients rather than landmarks. \emph{Expressivity} quantifies the intensity of expression activations. \emph{Diversity}, captures how many different expression types appear across time (i.e., variability)~\citep{ParishMorris2018}.

\begin{lstlisting}[style=bitboxstyle, language=Python]
# compute expressivity statistics
exp_stats = expressivity(exp_global, scales=6)

# compute diversity metrics at specified temporal scales
div_scores = diversity(exp_global, scales=[0.5, 1, 1.5, 2])
\end{lstlisting}

\begin{figure}[h]
    \centering
    \includegraphics[width=1.0\linewidth]{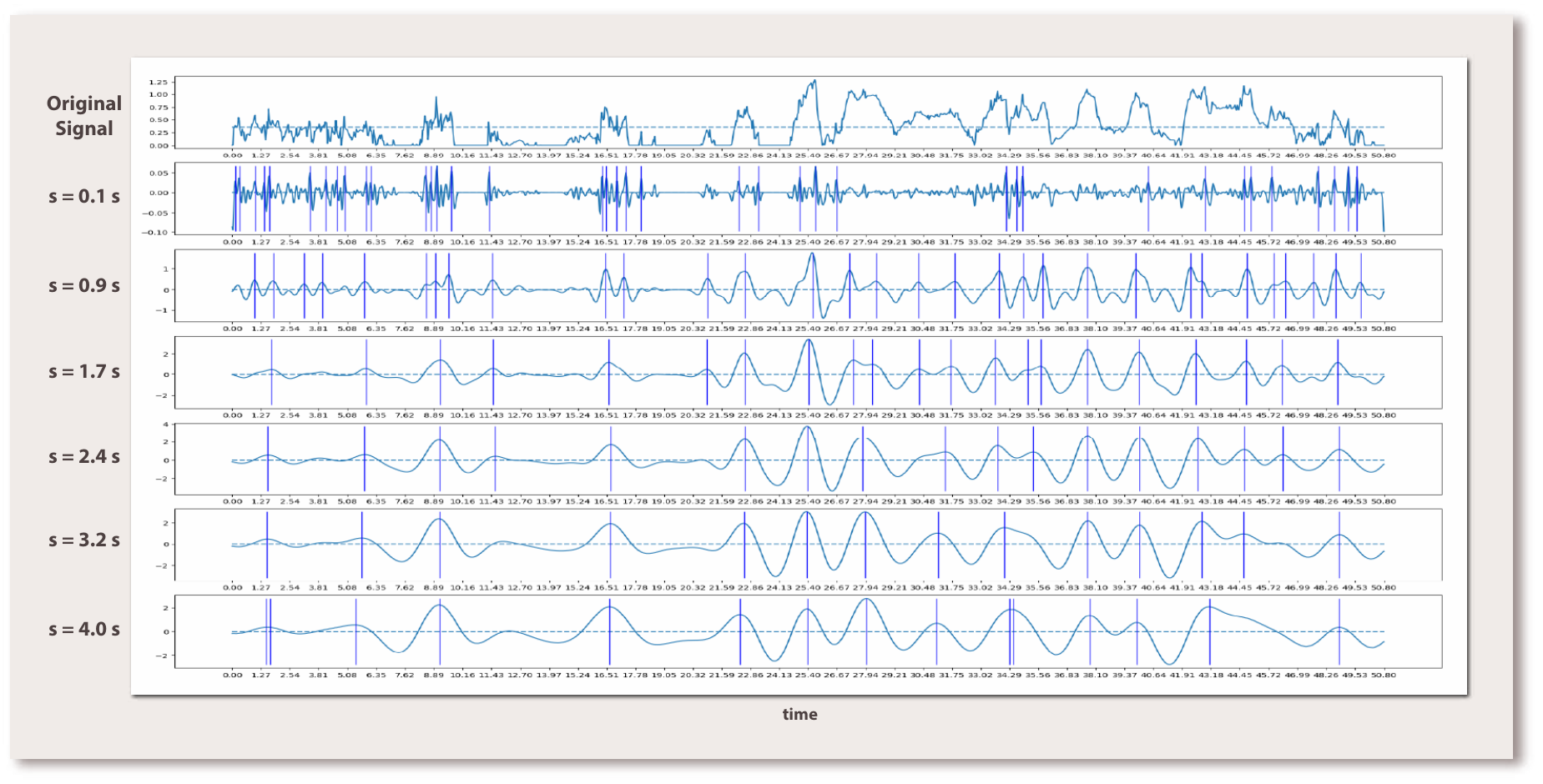}
    \caption{Decomposition of an expression signal into multiple components at different temporal scales ($s$). For each scale, automatically detected peaks that correspond to expression events occurring at different speeds are shown as blue vertical lines.} 
    \label{fig:multiscale}
\end{figure}

Both measures can be calculated at one (\texttt{scales=None}) or multiple temporal scales, where each scale represents a time window (in seconds) over which expression activity is analyzed. Shorter scales (0.1–0.5 s) capture rapid, fine-grained muscle activations, while longer scales (1–3 s) reflect sustained behavior such as gradual smiles, frowns, or prolonged emotional responses. Figure~\ref{fig:multiscale} illustrates how a single expression signal can be decomposed into several temporal scales, each revealing distinct patterns of expressive dynamics. Selecting appropriate scales depends on the nature of your data and research questions; when uncertain, analyzing multiple scales often provides a richer understanding of expression behavior.

Social-dynamics measurements, \emph{imitation} and \emph{coordination}, accept any combination of head pose, landmarks, or expression signals. Unlike the previous measures, these functions operate on two sets of signals. These may belong to two different people in a social interaction or to the same person (e.g., within-person coordination of facial expressions).

\begin{lstlisting}[style=bitboxstyle, language=Python]
# imitation: participant follows a reference model
corr_mean, corr_std, corr_lag = imitation(
    participant, reference, width=1.1, step=0.5, fps=30
)

# coordination: mutual synchrony between two partners
corr_mean, corr_std, corr_lag = coordination(
    participant_a, participant_b, width=1.1, step=0.5, fps=30
)
\end{lstlisting}

Both measures compute a windowed, lagged cross-correlation across all pairs of signals. The distinction lies in how lags are interpreted. Imitation assumes a directional relationship: the participant follows the reference, so lags are defined relative to the second signal and restricted to one direction (the imitator may lag behind but cannot precede the model). Coordination, by contrast, allows bidirectional lags, enabling each signal to lead or follow the other, which is essential for studying naturalistic interpersonal synchrony. If directional constraints are not desired, imitation can be run with \texttt{causality=False}, though lags will still be referenced to the second signal; for symmetric analysis, coordination is the appropriate choice.

Window width (\texttt{width}) and step size (\texttt{step}) are specified in seconds, so the frame rate (\texttt{fps}) must correspond to the sampling rate of the signal. Bitbox automatically identifies the optimal lag between signals within each window, allowing researchers to quantify how coupling, mimicry, or synchrony evolves over time.

\section{Visualization}
Bitbox provides a set of advanced visualization tools tailored for facial and body analysis. These tools are built on top of Plotly, an open-source visualization library that enables interactive, web-based graphics. We adapted Plotly’s capabilities to create interfaces optimized for behavioral imaging—supporting smooth exploration of movement signals, expression dynamics, and 3D facial structure. This allows researchers to inspect processed results in a way that is both intuitive and scientifically informative.

\begin{figure}[h]
    \centering
    \includegraphics[width=1.0\linewidth]{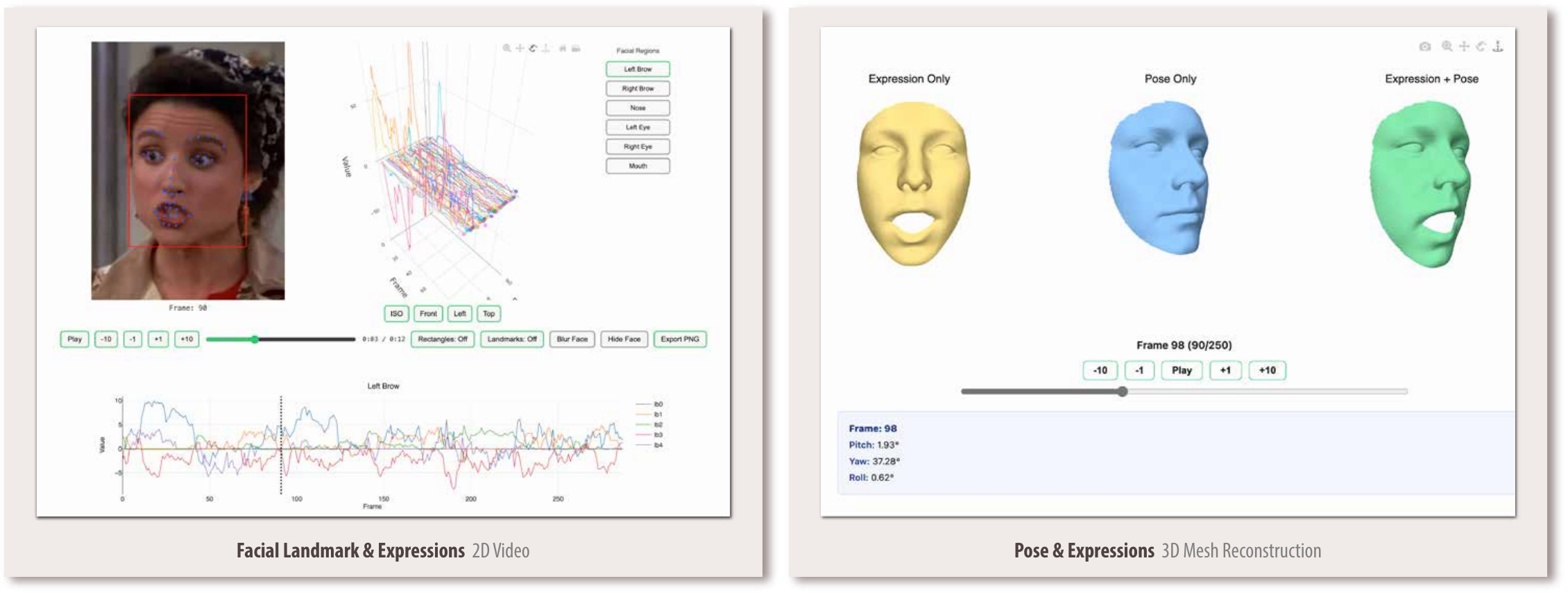}
    \caption{Interactive visualizations from Bitbox for 2D facial videos (left) and corresponding 3D reconstructions (right). Visuals can be exported as PNGs for publications or presentations. For videos, faces can be blurred or fully hidden to protect personally identifiable information (PII).}
    \label{fig:visuals}
\end{figure}

Bitbox can display face rectangles, 2D and 3D facial landmarks, head-pose trajectories, expression time-series, and reconstructed 3D facial meshes. Several examples of these visual outputs are shown in Fig.~\ref{fig:visuals}. The plotting interface is straightforward and flexible:

\begin{lstlisting}[style=bitboxstyle, language=Python]
# visualize landmarks at diverse head poses
processor.plot(lands, pose=pose)

# visualize landmarks with face rectangles overlayed
processor.plot(lands, overlay=[rects], video=True)

# visualize expression activations with overlays
processor.plot(exp_global, overlay=[rects, lands], video=True)
\end{lstlisting}

Each visualization is exported as an HTML/JavaScript file in the user-specified output directory. These files can be opened directly in any modern web browser. When loading a visualization, a brief delay is expected: the embedded JavaScript dynamically renders frames and builds interactive elements in the browser. Runtime depends on the user’s hardware and browser, but within a few seconds the visualization becomes fully interactive—allowing zooming, panning, rotating, and detailed inspection of the recorded behavior.

Bitbox’s visualization suite is under active development. We are expanding these tools to support additional modalities and new forms of interactive inspection, ensuring that researchers have access to high-quality, task-specific visualization environments as new needs emerge in behavioral and clinical science.

\section{Data and Metadata Management}
\label{sec:metadata}
Running advanced backend processors to produce output files might take some time, usually a few minutes per video file, depending on specific hardware requirements. To enhance analysis efficiency and provide a versioning system, Bitbox includes an integrated simple file caching mechanism.

Each time a processor is run, Bitbox checks whether the output files and their metadata, which detail the last execution, already exist in the specified output directory. This metadata is stored as \texttt{.json} files. If the files and metadata are found, Bitbox verifies if the time elapsed since their last creation is within the retention period (default is 6 months). If it is, Bitbox uses the existing files, avoiding the need to recreate them. This process significantly saves time. The user can easily adjust the retention time according to their requirements.

\begin{lstlisting}[style=bitboxstyle, language=Python]
# the retention period can be set using natural language
# 1 year, 3 minutes, 7 seconds etc.
processor.cache.change_retention_period('1 year')
\end{lstlisting}

When a backend generates a file and saves it to the disk, an accompanying \texttt{.json} file, named identically, is created by Bitbox to track the details of the most recent execution. An example metadata is given in Code~\ref{code:metadata}

\begin{lstlisting}[style=bitboxstyle,
caption={Example metadata stored in .json files},
label={code:metadata}]
"backend": "3DI",
"morphable_model": "BFMmm-19830",
"camera": 30,
"landmark": "global4",
"fast": false,
"local_bases": "0.0.1.F591-cd-K32d",
"input_hash": "4e31c4610ad3641ed651394827520add6d87efc5618c162",
"cmd": "CUDA_VISIBLE_DEVICES=1 docker run --rm --gpus device=1 -v /home/test/bitbox/tutorials/data:/app/input -v /home/test/bitbox/tutorials/output:/app/output -w /app/3DI bitbox:cuda12 ./video_detect_landmarks /app/input/elaine.mp4 /app/output/elaine_rects.3DI /app/output/elaine_landmarks.3DI /app/3DI/configs/BFMmm-19830.cfg1.global4.txt > /dev/null",
"input": "/home/test/bitbox/tutorials/data/elaine.mp4",
"output": "/home/test/bitbox/tutorials/output",
"time": "2025-07-18 12:33:50"
\end{lstlisting}

\section{Open Science and Reproducibility}
Reproducibility is a core priority in the design of Bitbox. As described earlier in Section~\ref{sec:metadata}, every file produced by Bitbox is accompanied by automatically generated metadata. This metadata includes the full command used to run the backend processor, the exact parameters passed to the executable, runtime information, and relevant system details. By storing this information alongside the outputs, Bitbox ensures that analyses can be reconstructed precisely, shared across research groups, and audited long after data collection or processing has taken place. This creates a transparent workflow in which behavioral measurements are not isolated results but fully traceable computational products.

At present, Bitbox does not implement full file versioning. Each output is accompanied by creation-time metadata, but repeated processing will overwrite previously generated files. However, the current metadata structure and internal caching system were designed with future versioning in mind. We are actively exploring more complete solutions, including potential integration with DataLad~\cite{Halchenko2021}, which offers a robust, research-oriented framework for dataset version control. Such an integration would make it possible to track the full evolution of processed datasets, facilitating multi-lab collaboration, preregistered analyses, and long-term provenance tracking.

To support transparency and proper scholarly citation, all Bitbox backends include a \texttt{processor.citation()} function. Calling this function produces a ready-to-use citation block for the specific backend, ensuring that downstream publications accurately acknowledge the tools and models used for behavioral signal extraction. An example is shown in Code~\ref{code:citation}. This feature promotes responsible tool usage and helps maintain clarity in the growing methodological ecosystem surrounding computational behavior analysis.

Our ongoing open-science efforts include expanding provenance tracking, supporting standardized behavioral-data formats, building automated quality-control reports, and developing tools that assist researchers in preparing reproducible workflows and supplemental materials directly from Bitbox output. These initiatives are actively under development as we work toward a fully reproducible and interoperable behavioral-measurement platform.

Finally, because Bitbox is modular and fully open-source, extensions and modifications can be added with minimal friction. We welcome contributions, suggestions, and community-driven improvements. As computational behavior analysis continues to evolve, we expect Bitbox to grow in parallel through collective effort, providing a shared foundation for transparent, scalable, and reproducible behavioral research.

\begin{lstlisting}[style=bitboxstyle,
caption={Example citation block provided by \texttt{processor.citation()}. It provides both the text block that can be directly used in manuscripts as well as the cited references.},
label={code:citation}]
The dataset was processed using Bitbox version 2025.10.dev2 for facial behavior analysis. Facial modeling was performed with the 3DI [1], configured with the Basel Face Model (BFM) 2009 [2]. For 3DI, the camera field of view was set to 30 degrees, and the iBUG-51 landmark template [3] was used for landmark definition.
               
[INCLUDE THE FOLLOWING IF YOU USED LOCAL EXPRESSION COEFFICIENTS]
Localized expression coefficients were computed using Facial Basis [4]. 

[1] Sariyanidi E, Zampella CJ, Schultz RT, Tunc B (2024). Inequality-Constrained 3D Morphable Face Model Fitting. IEEE Transactions on Pattern Analysis and Machine Intelligence, 46(2), 1305-1318. https://doi.org/10.1109/TPAMI.2023.3334948
[2] Paysan P, Knothe R, Amberg B, Romdhani S, Vetter T (2099). A 3D Face Model for Pose and Illumination Invariant Face Recognition. In Proceedings of the IEEE International Conference on Advanced Video and Signal based Surveillance (AVSS), 296-301. https://doi.org/10.1109/AVSS.2009.58
[3] Sariyanidi E, Zampella CJ, Schultz RT, Tunc B (2020). Can facial pose and expression be separated with weak perspective camera? In Proceedings of the IEEE/CVF Conference on Computer Vision and Pattern Recognition (CVPR), 7173-7182. https://doi.org/10.1109/CVPR42600.2020.00720
[4] Sariyanidi E, Yankowitz L, Schultz RT, Herrington JD, Tunc B, Cohn J (2025). Beyond FACS: Data-driven facial expression dictionaries, with application to predicting autism. In Proceedings of the IEEE International Conference on Automatic Face and Gesture Recognition (FG), 19, 1-10. https://doi.org/10.1109/fg61629.2025.11099288
\end{lstlisting}

\bibliography{references}% common bib file

\end{document}